\documentclass[sigconf]{acmart}

\usepackage{algorithm}      
\usepackage{algpseudocode}  
\usepackage{bm}
\usepackage{multirow}
\usepackage{float}

\AtBeginDocument{%
  }

\copyrightyear{2025} 
\acmYear{2025} 
\setcopyright{acmlicensed}
\acmConference[MM '25]{Proceedings of the 33rd ACM International Conference on Multimedia}{October 27--31, 2025}{Dublin, Ireland}
\acmBooktitle{Proceedings of the 33rd ACM International Conference on Multimedia (MM '25), October 27--31, 2025, Dublin, Ireland}
\acmDOI{10.1145/3746027.3755219}
\acmISBN{979-8-4007-2035-2/2025/10}

\begin{document}

\title{Exploring Multimodal Prompts For
Unsupervised Continuous Anomaly Detection}

\author{Mingle Zhou}
\orcid{0000-0003-4911-276X}
\affiliation{
  \institution{Key Laboratory of Computing Power Network and Information Security, Ministry of Education, Shandong Computer Science Center (National Supercomputer Center in Jinan), Qilu University of Technology (Shandong Academy of Sciences)}
  \city{Jinan}
  \country{China}
}
\email{zhouml@qlu.edu.cn}

\author{Jiahui Liu}
\orcid{0009-0003-4961-031X}
\affiliation{
  \institution{Key Laboratory of Computing Power Network and Information Security, Ministry of Education, Shandong Computer Science Center (National Supercomputer Center in Jinan), Qilu University of Technology (Shandong Academy of Sciences)}
  \city{Jinan}
  \country{China}
}
\email{10431240259@stu.qlu.edu.cn}

\author{Jin Wan}
\orcid{0000-0001-9245-0110}
\affiliation{
  \institution{Shandong Provincial Key Laboratory of Computing Power Internet and Service Computing, Shandong Fundamental Research Center for Computer Science}
  \city{Jinan}
  \country{China}
}
\email{wanj@qlu.edu.cn}

\author{Gang Li}
\orcid{0000-0002-7896-4833}
\affiliation{
  \institution{Key Laboratory of Computing Power Network and Information Security, Ministry of Education, Shandong Computer Science Center (National Supercomputer Center in Jinan), Qilu University of Technology (Shandong Academy of Sciences)}
  \city{Jinan}
  \country{China}
}
\affiliation{
  \institution{Faculty of Data Science, City University of Macau}
  \city{Macau}
  \country{China}
}
\email{lig@qlu.edu.cn}
 
\author{Min Li}
\orcid{0000-0002-0507-5576}
\authornote{Corresponding author}
\affiliation{
  \institution{Faculty of Data Science, City University of Macau}
  \city{Macau}
  \country{China}
}
\affiliation{
  \institution{Key Laboratory of Computing Power Network and Information Security, Ministry of Education, Shandong Computer Science Center (National Supercomputer Center in Jinan), Qilu University of Technology (Shandong Academy of Sciences)}
  \city{Jinan}
  \country{China}
}
\email{limin@qlu.edu.cn }

\renewcommand{\shortauthors}{Mingle Zhou et al.}

\begin{abstract}
Unsupervised Continuous Anomaly Detection (UCAD) is gaining attention for effectively addressing the catastrophic forgetting and heavy computational burden issues in traditional Unsupervised Anomaly Detection (UAD).           However, existing UCAD approaches that rely solely on visual information are insufficient to capture the manifold of normality in complex scenes, thereby impeding further gains in anomaly detection accuracy.       To overcome this limitation, we propose an unsupervised continual anomaly detection framework grounded in multimodal prompting.       Specifically, we introduce a Continual Multimodal Prompt Memory Bank (CMPMB) that progressively distills and retains prototypical normal patterns from both visual and textual domains across consecutive tasks, yielding a richer representation of normality.          Furthermore, we devise a Defect-Semantic-Guided Adaptive Fusion Mechanism (DSG-AFM) that integrates an Adaptive Normalization Module (ANM) with a Dynamic Fusion Strategy (DFS) to jointly enhance detection accuracy and adversarial robustness.     Benchmark experiments on MVTec AD and VisA datasets show that our approach achieves state-of-the-art (SOTA) performance on image-level AUROC and pixel-level AUPR metrics.
\end{abstract}




\maketitle
\section{Introduction}
\begin{figure} 
\centering 
\includegraphics[width=1\linewidth]{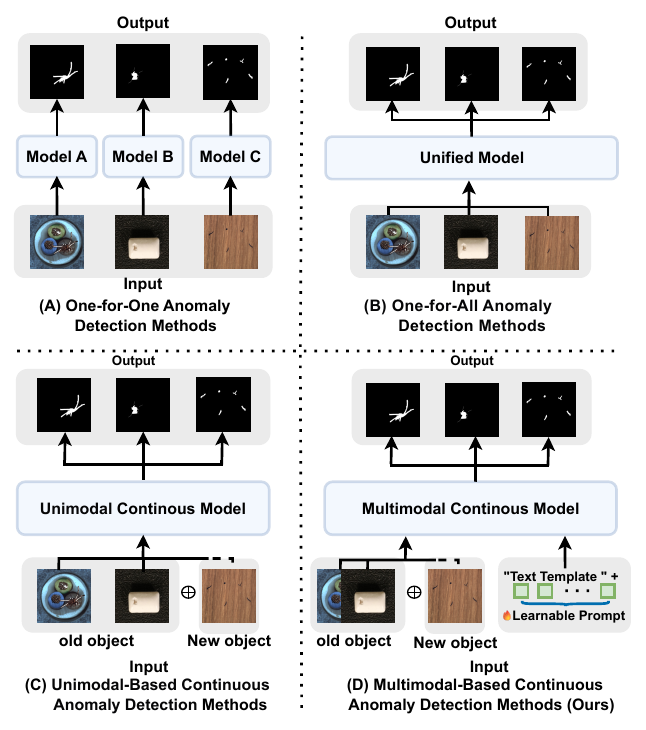} 
\Description{Panels A–D contrast four anomaly-detection frameworks: one-vs-one, unified multiclass, continual-learning integration, and our proposed multimodal continual approach.}
\caption{Different frameworks in anomaly detection. (A) illustrates the prevalent One-for-One Anomaly Detection paradigm \cite{4,20,reiss2021panda}. 
(B) depicts the unified multiclass model, streamlining the anomaly detection process across multiple classes \cite{54}. (C) showcases the integration of continual learning into anomaly detection \cite{8}. (D) represents our proposed multimodal continual anomaly detection model. }
\label{Fig.1} 
\end{figure}
In the rapid advancement of industrial automation, anomaly detection has emerged as an indispensable key technology in industrial manufacturing. It plays a crucial role in ensuring product quality, enhancing production efficiency, and reducing costs. Traditional supervised anomaly detection methods \cite{1,2,3} typically rely on extensive manually annotated data for training. However, acquiring large-scale labeled data in real-world scenarios is both costly and time-consuming. Consequently, unsupervised anomaly detection (UAD) has become an ideal solution for addressing anomaly detection challenges in the absence of labeled data. UAD can effectively identify abnormal data that significantly deviates from normal samples by learning the characteristics of normal data alone. Nevertheless, current mainstream UAD approaches face various challenges. The one-for-one anomaly detection paradigm \cite{4,20,reiss2021panda} achieves UAD exception detection by training separate models for each class. In this setup, sequential model learning imposes a heavy computational burden as the number of classes increases. Other methods \cite{6,7} focus on training uniform anomaly detection models across multiple classes. However, in practice, sequential training is often required, making it impractical to train all classes simultaneously. Moreover, these methods cannot effectively retain previously learned knowledge in the face of frequent product changes, leading to catastrophic forgetting.

To address these issues, Continuous Learning (CL) offers an effective framework for retaining historical knowledge while learning new tasks, thereby mitigating the catastrophic forgetting problem experienced by unified models such as those in \cite{6} under conditions of frequent product changes. Although recent work by Liu et al. \cite{8} demonstrates the efficacy of CL in unsupervised anomaly detection, their approach relies solely on visual information for anomaly detection, neglecting the complementary potential of multimodal data. This limitation has become a bottleneck to further performance improvements. Given the proven effectiveness of multimodal information in anomaly detection tasks, it is essential to explore the application of multimodal prompts in Unsupervised Continuous Anomaly Detection (UCAD) to overcome its current performance bottleneck.

In this study, we propose an unsupervised continuous anomaly detection framework based on multimodal prompts.  Our core innovation lies in the introduction of the Continuous Multimodal Cue Memory (CMPMB), which integrates learnable textual and visual prompts to refine normal feature representation while enabling lifelong knowledge retention progressively. CMPMB incorporates task-specific identification keys for rapid task recognition, adaptive multimodal prompts for cross-modal alignment, and a general feature repository for mitigating forgetting. To enhance the integration of multimodal information and further improve the model's segmentation performance, we design a defect semantics-guided adaptive fusion mechanism (DSG-AFM), which combines an adaptive normalization mechanism (ANM) and a dynamic fusion strategy (DFS) for context-aware multimodal feature fusion. Our contributions are summarized as follows:

\begin{itemize}
\item  We propose a multimodal continuous anomaly detection method that integrates text and visual prompts by constructing a continuous multimodal prompt memory Bank (CMPMB) to optimize normal feature representation and significantly improve cross-task generalization.
\item  We propose a defect Semantics-guided fusion mechanism (DSG-AFM) that achieves effective fusion of multi-modal information by dynamically balancing multi-modal features through adaptive normalization and fusion strategy.
\item  Our proposed method outperforms the previous state-of-the-art (SOTA) AD method by achieving a 4.4\% higher accuracy in detection and a 14.8\% higher accuracy in segmentation.
\end{itemize}

\section{Related Work}
\label{re_work}
\subsection{Multimodal Anomaly Detection}
Multimodal cue learning has emerged as a critical advancement in anomaly detection, particularly through the application of cross-modal feature alignment techniques \cite{7,9,10}. Notable contributions include SAA+ \cite{11}, which employs domain-expert multimodal prompts to achieve zero-shot anomaly segmentation with improved localization accuracy, and MMRD \cite{12}, which proposes a multimodal dedistillation framework for robust multi-class anomaly detection. IPDN \cite{13} mitigates cross-modal semantic ambiguity via multi-view semantic embedding and an instant perception decoder. Recent advancements \cite{14,15} further highlight the adaptability of multimodal prompts through flexible architectural designs.

\subsection{Unsupervised Anomaly Detection}
In unsupervised anomaly detection, methodologies are primarily categorized into feature-embedding and reconstruction-based approaches. Feature-embedding methods include: 1) Teacher-student frameworks \cite{16,17} for knowledge distillation; 2) Single-class classification methods \cite{18,19,20}; 3) Pre-trained feature space mapping \cite{21,22,23}; and 4) Memory-enhanced architectures \cite{24,25}. Reconstruction-based approaches utilize various network structures, such as autoencoders \cite{26,27}, GANs \cite{28}, Vision Transformers \cite{29}, and diffusion models \cite{30}. Despite significant progress, these methods predominantly focus on single-modal inputs, neglecting the potential of multimodal data for feature complementarity.

\subsection{Continuous Anomaly Detection}
Continuous anomaly detection (CAD) confronts dual challenges in incremental knowledge retention and fine-grained anomaly segmentation. Traditional methods \cite{zhang2023iddm,31,32} mainly address intra-class anomalies, while recent work UCAD \cite{8} introduces structural contrastive learning but remains limited by unimodal visual inputs. To overcome these limitations, we propose a multimodal prompt-based CAD framework incorporating two key innovations: 1) Continuous Multimodal Prompt Memory Bank (CMPMB) for preserving cross-modal knowledge, and 2) Defect Semantics-Guided Adaptive Fusion Mechanism (DSG-AFM) for pixel-level anomaly localization. This dual-level architecture enhances both interclass and intraclass detection at the classification level, enabling precise anomaly segmentation through dynamic multimodal fusion and effectively addressing the performance bottlenecks of existing methods.

\begin{figure*} 
\centering 
\includegraphics[width=1\linewidth]{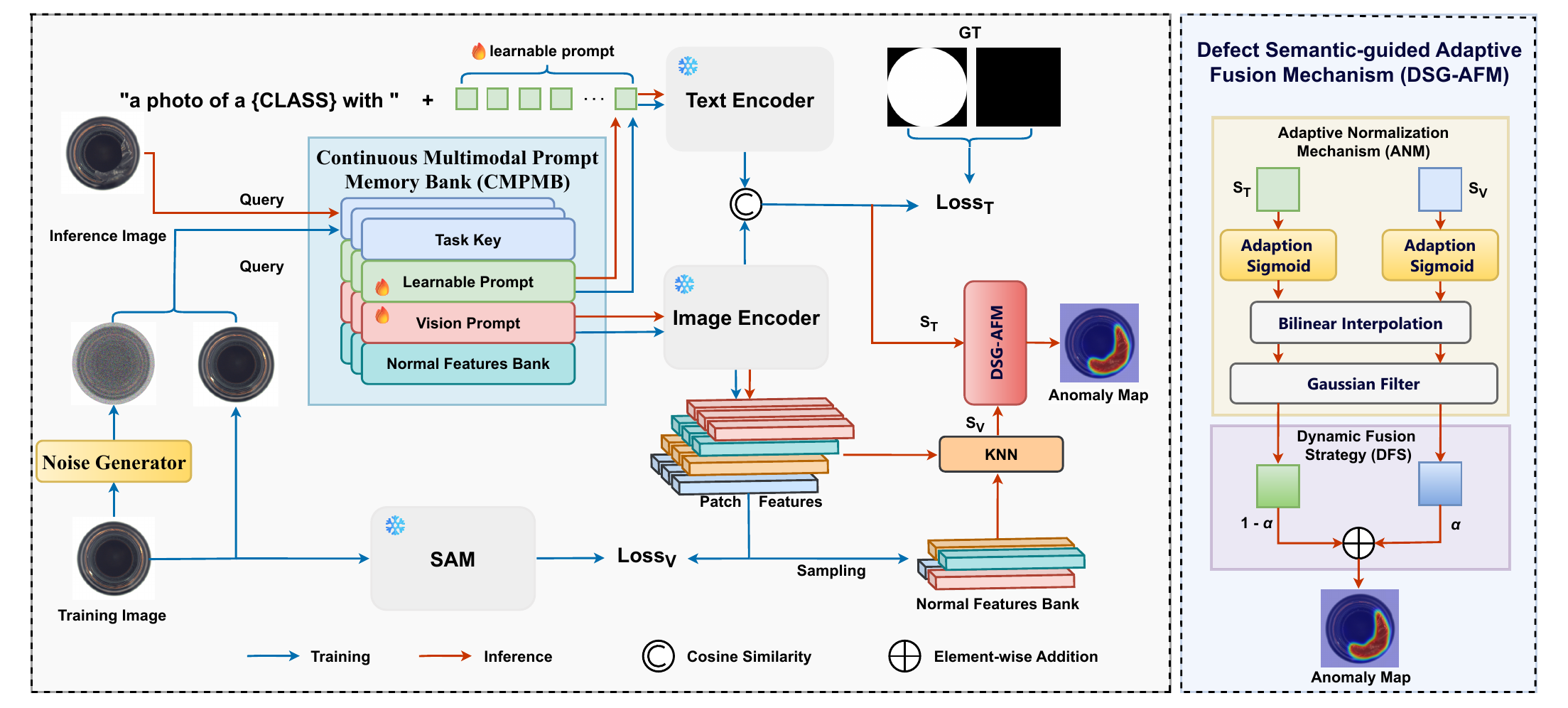} 
\Description{The architecture of our model consists of two core components.}
\caption{The architecture of our model consists of two core components. The Continuous Multimodal Prompts Memory Block (CMPMB) and the Defect Semantics-Guided Adaptive Fusion Mechanism (DSG-AFM).     The CMPMB endows the model with continuous learning ability and facilitates unsupervised anomaly detection in continuous scenarios.     The DSG-AFM further enhances anomaly detection accuracy by refining the anomaly map based on defect semantics and a dynamic normalization mechanism.} 

\label{Fig.2} 
\end{figure*}

\section{Methodology}
\label{sec:method}
\subsection{Overview}
This paper introduces a continuous unsupervised anomaly detection method leveraging multimodal prompts. As illustrated in \ref{Fig.2}, our approach incorporates a Continuous Multimodal Prompt Memory Bank (CMPMB) that integrates learnable textual and visual prompts to progressively refine normal feature representations and enable continuous task knowledge updates. The CMPMB comprises task-specific keys for rapid task identification, adaptive multimodal prompts for cross-modal alignment, and a normal feature repository to mitigate catastrophic forgetting. The unified architecture synergistically balances plasticity and stability in lifelong anomaly detection. To enhance anomaly localization, we propose a defect semantics-guided adaptive fusion mechanism (DSG-AFM) with an adaptive normalization module (ANM) and dynamic fusion strategy (DFS) for context-aware multimodal feature fusion. This framework significantly improves the accuracy and robustness of anomaly detection.
\subsection{Continuous Multimodal Prompt Memory Bank}
In an industrial automation environment, tasks are typically executed in a sequential pipeline. Although traditional multi-category anomaly detection methods can identify multiple categories simultaneously, they require all categories to be available for training at once, which is impractical in real-world applications. To address unsupervised multimodal anomaly detection in continuous learning scenarios, we propose the integration of learnable text prompts and refined visual prompts to construct a Continuous Multimodal Prompt Memory Bank (CMPMB). The CMPMB integrates task identification ($K$), learnable text prompts ($P^{T}$), refined visual prompts ($P^{V}$), and a normal feature library (F) through the quadruple $M = (K, P^{T}, P^{V}, F)$, achieving collaborative optimization of modal features and efficient knowledge accumulation. For each task t, we construct a quadruple $M_t = (K_t, P_t^{T}, P_t^{V}, F_t)$ to implement unsupervised persistent anomaly detection.

During the task adaptation phase, the task identifier Kt filters key embedders from patch-level features extracted from a frozen pre-trained visual backbone network using farthest point sampling (FPS) \cite{eldar1997farthest}. This process builds low-dimensional, high-discriminative task representations and stores them in Mt for task-agnostic predictions. Since task identities may not always be available in real-world scenarios, task-agnostic predictions are essential. In the task adaptation stage, for the image set $x \in \mathbb{R}^{N_t \times C \times H \times W}$ of task t, where $N_t$ is the number of images in task t, and $C \times H \times W$ represents the size of a single image, we extract task-related identity information 
$
    K_t = f_j(x), K_t \in \mathbb{R}^{N_t \times N_p \times C}
$ and $N_p$ is the number of patch features. $C$ denotes the feature dimension. In this study, we use the fifth layer of features because mid-tier features provide both contextual and semantic information for task identifiers \cite{8}. Define the set of tasks $T = \{t_1, t_2, ..., t_n\}$, then the identity of task $K$ can be expressed as:
\begin{align}
K & = \{K_{t_{1}}', K_{t_{2}}', \ldots, K_{t_{n}}'\}, K \in \mathbb{R} ^{N_f \times C} \\ 
  & = \{\text{FPS}(K_{t_{1}}), \text{FPS}(K_{t_{2}}), \ldots, \text{FPS}(K_{t_{n}})\}, 
\end{align} where n is the number of tasks, FPS is farthest point sampling. In the task reasoning stage, for an image $x \in \mathbb{R}^{C \times H \times W}$, we extract identity information $k'$ through the frozen pre-trained visual backbone network and calculate the highest similarity between $k'$ and the task identifier $k$ to identify the task identity. Additionally, we use the fine-tuned pre-trained visual backbone network to extract normal features and form a normal feature library. Similar to task identification $K$, we employ coreset sampling (CSS) \cite{24} to compress the extracted patch-level features, resulting in the final normal feature library $F$. Define the normal feature set $F_{t}^{s} = \{f_{1}^{s}, f_{2}^{s}, \ldots, f_{m}^{s}\}$, where $m$ is the number of normal patch features. Then $F$ can be expressed as follows: \begin{align}
    F & = \{F_{t_{1}}', F_{t_{2}}', \ldots, F_{t_{n}}'\} \\
      & = \{\text{CSS}(F_{t_{1}}), \text{CSS}(F_{t_{2}}), \ldots, \text{CSS}(F_{t_{n}})\},  \\
      \text{CSS}(F_{t_{n}}) & = \{f_{1}^{c}, f_{2}^{c}, \ldots, f_{k}^{c}\}, k \leq m  \\
       & =  \arg \min_{F_{t}^{c} \subset F_{t}^{s}} \max_{f_{j} \in F_{t}^{s}} \min_{f_{i} \in F_{t}^{c}} \| f_{j} - f_{i} \|_2, 
\end{align}
where CSS is coreset sampling, $F_{t_n} \in \mathbb{R}^{Ng \times C}$, $N_{g}$ is the number of patch features in the core set of the normal feature $F_{t}^{s}$, and $F_{t}^{c}$ is the space of the core set of the normal feature $F_{t}^{s}$ extracted during the task adaptation phase.
The multimodal prompt mechanism is pivotal for achieving efficient and continuous unsupervised anomaly detection. This mechanism facilitates cross-modal feature alignment and task-adaptive guidance through the collaborative optimization of textual and visual prompts. Specifically, for text prompts, we draw inspiration from the CLIP \cite{37} multimodal alignment framework by introducing a learnable prompt template S in the form "$a$ $photo$ $of$ $a$ $[class]$ $with$ $[P_{t}^{T}]$", where $[class]$ denotes a specific task category (e.g., "$Bottle$"), and $P_{t}^{T} \in \mathbb{R}^{N_{l} \times C}$ represents a set of learnable vectors. In this study, $N_{l}$ is set to 5, with each $p_i(i \in N_l)$ initialized from a standard normal distribution $\mathcal{N}(0,1)$, to encode the semantic distribution of normal samples. During the task adaptation phase, the learnable text prompt Pt is optimized via contrastive learning using noise-enhanced samples. To optimize the learnable text prompts, we employ the Mean Squared Error (MSE) as the loss function:
\begin{equation}
    Loss_{T} = \frac{1}{N} \sum_{i=1}^N (x_i - y_i)^2,
\end{equation}
where $x_i$ denotes the output of the model, $y_i$ represents the corresponding label value, and $N$ is the total number of samples. This approach enables the model to align the feature distribution of normal samples with the text prompt without requiring manual labeling, thereby facilitating the learning of normal patterns. For an image $x \in \mathbb{R}^{C \times H \times W}$, Gaussian noise is added to generate an auxiliary image:
\begin{align}
    x_{noise} &= NoiseGenerator(x) \\
    &= x + \mathcal{N}(0, \sigma^2),
\end{align}
where $\mathcal{N}(0, \sigma^2) \in \mathbb{R}^{C \times H \times W}$, and in this paper, $\sigma$ is set to 1. 
Inspired by Li et al. \cite{8}, we introduce a hierarchical refinement visual prompt $P^V_t$ to guide unsupervised anomaly detection. The task-aware visual prompts $P_t^V \in \mathbb{R}^{N_v \times l \times C}$ consist of $N_v$ learnable vectors with dimension $l \times C$, where $l$ denotes the prompt length (set to 2 in our implementation) and $C$ represents the channel dimension. These visual prompts are initialized through uniform distribution $P^V_t \sim Uniform(0,1)$. During the task adaptation and inference stages, we employ prefix tuning by integrating a visual prompt $\hat{P^V_{t}} \in \mathbb{R}^{l \times C}$ into the input features of each layer in the pre-trained visual backbone network. This process enables layer-wise propagation of task-specific information, formally expressed as:
\begin{equation}
    x_i = f_i(x_{i-1} + P^V_{i}), P^V_{i} \in \mathbb{R}^{l \times C},
\end{equation}
where $x_i$ denotes the output features from the $i$-th layer of the backbone network, $x_{i-1}$ represents the features from the $(i-1)$-th layer, and $P^{V}_{i}$ indicates the visual prompt specifically learned for layer $i$ to facilitate fine-grained task adaptation. The detailed implementation of this visual prompt tuning process $f_i(x_{i-1} + v_i)$ is presented in Algorithm \ref{alg:prompt_attn}.
To enhance feature compactness and optimize storage efficiency, we introduce a structured contrastive loss \cite{8} for visual prompt optimization. The loss function is formulated as:
\begin{align}
    Loss_{V}^{pos} &= \sum_{i,p=1}^{H} \sum_{j,q=1}^{W} \cos(F_{ij}, G_{pq}), \quad (F_{ij} = G_{pq}),  \\
    Loss_{V}^{neg} &= \sum_{i,p=1}^{H} \sum_{j,q=1}^{W} \cos(F_{ij}, G_{pq}), \quad (F_{ij} \neq G_{pq}),  \\
    Loss_{V} &= \lambda_{\alpha} Loss_{V}^{neg} - \lambda_{\beta} Loss_{V}^{pos},
\end{align}
where $F_{ij}$ represents the embedding of feature $F$ at the position of $(i, j)$, and the shape is $F_{ij} \in \mathbb{R}^{1 \times 1 \times c}$, while $G_{pq}$ represents the embedding of feature $F_{ij}$ at the corresponding position in the segmentation result generated by SAM. Similarly, $G_{pq} \in \mathbb{R}^{1 \times 1 \times c}$.
$\lambda_{\alpha}$ and $\lambda_{\beta}$ both have the value of one.

CMPMB achieves efficient task knowledge accumulation and migration through continuous iterative optimization of multimodal prompts. Textual prompts focus on semantic-level normal mode modeling, visual prompts strengthen local feature adaptation, and the two achieve cross-modal interactions by sharing task identifiers to form complementary feature representations. Experiments show that the design significantly improves the robustness and cross-task generalization ability of unsupervised anomaly detection.

\begin{algorithm}
\caption{Vision Prompt-Tuned Process}
\label{alg:prompt_attn}
\begin{algorithmic}[1]
\Require $\bm{x_{i-1}} \in \mathbb{R}^{B \times N \times C}$, $\bm{p^{V}_{i}} \in \mathbb{R}^{B \times l \times C}$, $h$
\Ensure Output feature $\bm{y}$
\State \textbf{Projection:} 
  \State $\bm{q}, \bm{k}, \bm{v} \gets \text{Linear}(\bm{x_{i-1}})$ \label{step:projection}
\State Reshape $\bm{q},\bm{k},\bm{v} \to [B, h, N, \frac{C}{h}]$
\If{$\bm{p_{i}} \neq \emptyset$} \label{step:prompt_fusion}
  \State Reshape $\bm{pi} \to [2, B, h, \frac{l}{2}, \frac{C}{h}]$ 
  \State Split $\bm{p_{i}}$ into $\bm{k}_{p_i}, \bm{v}_{p_i} \in \mathbb{R}^{B \times h \times \frac{l}{2} \times \frac{C}{h}}$ 
  \State $\bm{k} \gets \text{Concat}([\bm{k}_{p_i}, \bm{k}], \text{dim}=2)$ 
  \State $\bm{v} \gets \text{Concat}([\bm{v}_{p_i}, \bm{v}], \text{dim}=2)$ 
\EndIf
\State $\bm{A} \gets \text{Softmax}\left( \frac{\bm{q} \bm{k}^\top}{\sqrt{d}} \right)$ \label{step:softmax}
\State $\bm{A} \gets \text{Dropout}(\bm{A})$ \label{step:attn_drop}
\State $\bm{y} \gets \text{Linear}(\bm{A} \bm{v})$ 
\State Reshape $\bm{y} \to [B,N,C]$
\label{step:out_proj}
\State $\bm{y} \gets \text{Dropout}(\bm{y})$ 
\State \Return $\bm{y}$ \label{step:return}
\label{step:proj_drop}
\end{algorithmic}
\end{algorithm}

\subsection{Adaptive Fusion Mechanism Based On Defect Semantic Guidance}
To fully leverage the complementary advantages of multi-modal information, this paper introduces an Adaptive Fusion Mechanism based on Defect Semantic Guidance (DSG-AFM) to enhance the segmentation performance in anomaly detection. This mechanism employs a two-branch cooperative reasoning approach combined with a dynamic normalization strategy for efficient fusion of visual and textual features. The DSG-AFM framework comprises visual and textual branches: In the visual branch, a pre-trained Vision Transformer (ViT) is utilized to extract visual features, specifically using the fifth layer features for their semantically balanced and information-rich characteristics, which are crucial for task identification and anomaly detection. Initially, the sixth-layer features from the frozen vision backbone network calculate the highest similarity with the task identification stored in CMPMB to achieve rapid task identity matching. Subsequently, the task-associated visual prompt guidance model generates task-related feature output $F_{V}$ and calculates the nearest neighbor distance at the patch level with the normal feature library Ft stored in CMPMB, producing a preliminary patch-level anomaly score $S_{V}$.

\begin{table*}
\centering
\caption{Image-level AUROC $\uparrow$ and corresponding FM $\downarrow$ on MVTec AD dataset \cite{34} after training on the last subdataset. Note that * signifies the usage of a cache pool for rehearsal during training, which may not be possible in real applications. The best results are highlighted in \textbf{bold}, and suboptimal results are highlighted in \underline{underline}.}
\label{table:1}
\resizebox{\linewidth}{!}{ 
\begin{tabular}{l|ccccccccccccccc|cc}
\hline
\textbf{Methods} & \textbf{Bottle} & \textbf{cable} & \textbf{capsule} & \textbf{carpet} & \textbf{grid} & \textbf{hazelnut} & \textbf{leather} & \textbf{metal\_nut} & \textbf{pill} & \textbf{screw} & \textbf{tile} & \textbf{toothbrush} & \textbf{transistor} & \textbf{wood} & \textbf{zipper} & \textbf{average} & \textbf{avg FM} \\ \hline
CFA & 0.309 & 0.489 & 0.275 & 0.834 & 0.571 & 0.903 & \underline{0.935}& 0.464 & 0.528 & 0.528 & 0.763& 0.519 & 0.320 & 0.923 & 0.984 & 0.623& 0.361 \\ 
CSFlow & 0.129 & 0.420 & 0.363 & 0.978 & 0.602 & 0.269 & 0.906 & 0.220 & 0.263 & 0.434 & 0.697 & 0.569 & 0.432 & 0.802 & \textbf{0.997} & 0.539 & 0.426 \\ 
CutPaste & 0.111 & 0.422 & 0.373 & 0.198 & 0.214 & 0.578 & 0.007 & 0.517 & 0.371 & 0.356 & 0.112 & 0.158 & 0.340 & 0.150 & 0.775 & 0.312 & 0.510 \\ 
DRAEM & 0.793 & 0.411 & 0.517 & 0.537 & 0.799 & 0.524 & 0.480 & 0.422 & 0.452 & \textbf{1.000} & 0.548 & 0.625 & 0.307 & 0.517 & \underline{0.996}& 0.595 & 0.371 \\ 
FastFlow & 0.454 & 0.512 & 0.517 & 0.489 & 0.482 & 0.522 & 0.487 & 0.476 & 0.575 & 0.402 & 0.489 & 0.267 & 0.526 & 0.616 & 0.867 & 0.512 & 0.279 \\ 
FAVAE & 0.666 & 0.396 & 0.357 & 0.610 & 0.644 & 0.884 & 0.406 & 0.416 & 0.531 & 0.624 & 0.563 & 0.503 & 0.331 & 0.728 & 0.544 & 0.547 & 0.102 \\ 
PaDiM & 0.458 & 0.544 & 0.418 & 0.454 & 0.704 & 0.635 & 0.418 & 0.446 & 0.449 & 0.578 & 0.581 & 0.678 & 0.407 & 0.549 & 0.855 & 0.545 & 0.368 \\ 
PatchCore & 0.163 & 0.518 & 0.350 & 0.968 & 0.700 & 0.839 & 0.625 & 0.259 & 0.459 & 0.484 & 0.776 & 0.586 & 0.341 & 0.970 & 0.991 & 0.602 & 0.383 \\ 
RD4AD & 0.401 & 0.538 & 0.475 & 0.583 & 0.558 & 0.909 & 0.596 & 0.623 & 0.479 & 0.596 & 0.715 & 0.397 & 0.385 & 0.700 & 0.987 & 0.596 & 0.393 \\ 
SPADE & 0.302 & 0.444 & 0.525 & 0.529 & 0.460 & 0.410 & 0.577 & 0.592 & 0.484 & 0.514 & 0.881 & 0.386 & 0.622 & 0.897 & 0.949 & 0.571 & 0.285 \\ 
STPM & 0.329 & 0.539 & 0.610 & 0.462 & 0.569 & 0.540 & 0.740 & 0.456 & 0.523 & 0.753 & 0.736 & 0.375 & 0.450 & 0.779 & 0.783 & 0.576 & 0.325 \\ 
SimpleNet & 0.938 & 0.560 & 0.519 & 0.736 & 0.592 & 0.859 & 0.749 & 0.710 & 0.701 & 0.599 & 0.654 & 0.422 & 0.669 & 0.908 & \underline{0.996}& 0.708 & 0.211 \\ 
UniAD & 0.801 & 0.660 & 0.823 & 0.754 & 0.713 & 0.904 & 0.715 & 0.791 & 0.869 & 0.731 & 0.687 & 0.776 & 0.490 & 0.903 & \textbf{0.997} & 0.774 & 0.229 \\ \hline
DNE & 0.990 & 0.619 & 0.609 & 0.984 & \textbf{0.998}& 0.924 & \textbf{1.000} & \underline{0.989}& 0.671 & 0.588 & 0.980 & 0.933 & 0.877 & 0.930 & 0.958 & 0.870 & 0.116 \\ 
PatchCore* & 0.533 & 0.505 & 0.351 & 0.723 & \underline{0.959}& 0.854 & 0.456 & 0.511 & 0.626 & 0.748 & 0.600 & 0.427 & 0.900 & 0.974 & 0.669 & 0.229 & 0.318\\ 
UniAD* & \underline{0.997}& 0.701 & 0.765 & \underline{0.998}& 0.896 & 0.936 & \textbf{1.000} & 0.964 & \underline{0.895}& 0.554 & 0.989 & 0.928 & \textbf{0.966}& 0.982 & 0.987 & 0.904 & 0.076 \\ 
UCAD & \textbf{1.000} & \underline{0.751}& \underline{0.866}& 0.965 & 0.944 & \textbf{0.994}& \textbf{1.000} & 0.988 & 0.894 & 0.739 & \textbf{0.998}& \textbf{1.000} & 0.874 & \textbf{0.995} & 0.938 & \underline{0.930}& \underline{0.010}\\ \hline
Ours & \textbf{1.000} & \textbf{0.960}& \textbf{0.959}& \textbf{0.999}& \textbf{0.998}& \underline{0.992}& \textbf{1.000} & \textbf{0.996}& \textbf{0.962}& \underline{0.856}& \underline{0.996}& \underline{0.975}& \underline{0.946}& \underline{0.991}& 0.982& \textbf{0.974}& \textbf{0.009}\\ \hline
\end{tabular}
}
\end{table*}
\begin{table*}
\centering
\caption{Pixel-level AUPR $\uparrow$ and corresponding FM $\downarrow$ on MVTec AD dataset \cite{34} after training on the last subdataset.}
\label{table:2}
\resizebox{\linewidth}{!}{ 
\begin{tabular}{l|ccccccccccccccc|cc}
\hline
\textbf{Methods} & \textbf{Bottle} & \textbf{cable} & \textbf{capsule} & \textbf{carpet} & \textbf{grid} & \textbf{hazelnut} & \textbf{leather} & \textbf{metal\_nut} & \textbf{pill} & \textbf{screw} & \textbf{tile} & \textbf{toothbrush} & \textbf{transistor} & \textbf{wood} & \textbf{zipper} & \textbf{average} & \textbf{avg FM} \\ \hline
CFA & 0.068 & 0.056 & 0.050 & 0.271 & 0.004 & 0.341 & \underline{0.393}& 0.255 & 0.080 & 0.015 & 0.155 & 0.053 & 0.056 & 0.281 & 0.573 & 0.177 & 0.083 \\ 
DRAEM & 0.117 & 0.019 & 0.044 & 0.018 & 0.005 & 0.036 & 0.036 & 0.142 & 0.104 & 0.002 & 0.130 & 0.039 & 0.040 & 0.033 & \textbf{0.734} & 0.098 & 0.116 \\ 
FastFlow & 0.044 & 0.021 & 0.013 & 0.013 & 0.005 & 0.028 & 0.007 & 0.090 & 0.029 & 0.003 & 0.060 & 0.015 & 0.036 & 0.037 & 0.264 & 0.044 & 0.214 \\ 
FAVAE & 0.086 & 0.048 & 0.039 & 0.015 & 0.004 & 0.389 & 0.389 & 0.174 & 0.070 & 0.017 & 0.064 & 0.043 & 0.046 & 0.093 & 0.039 & 0.083 & 0.083 \\ 
PaDiM & 0.072 & 0.037 & 0.030 & 0.023 & 0.006 & 0.183 & 0.039 & 0.155 & 0.044 & 0.014 & 0.065 & 0.044 & 0.049 & 0.080 & 0.452 & 0.086 & 0.366 \\ 
PatchCore & 0.048 & 0.029 & 0.035 & 0.552 & 0.003 & 0.338 & 0.279 & 0.248 & 0.051 & 0.061 & 0.249 & 0.034 & 0.079 & 0.304 & 0.595 & 0.190 & 0.371 \\
RD4AD & 0.055 & 0.040 & 0.064 & 0.212 & 0.005 & 0.384 & 0.116 & 0.247 & 0.051 & 0.061 & 0.193 & 0.034 & 0.059 & 0.097 & 0.562 & 0.143 & 0.425 \\ 
SPADE & 0.122 & 0.052 & 0.044 & 0.117 & 0.004 & \underline{0.512}& 0.264 & 0.181 & 0.060 & 0.020 & 0.096 & 0.043 & 0.050 & 0.172 & 0.531 & 0.151 & 0.319 \\ 
STPM & 0.074 & 0.019 & 0.073 & 0.054 & 0.005 & 0.037 & 0.108 & 0.354 & 0.111 & 0.001 & 0.397 & 0.046 & 0.046 & 0.119 & 0.203 & 0.110 & 0.352 \\ 
SimpleNet & 0.108 & 0.045 & 0.029 & 0.018 & 0.004 & 0.029 & 0.006 & 0.227 & 0.077 & 0.004 & 0.082 & 0.046 & 0.049 & 0.037 & 0.139 & 0.060 & 0.069 \\
UniAD & 0.054 & 0.031 & 0.022 & 0.047 & 0.007 & 0.189 & 0.053 & 0.110 & 0.034 & 0.008 & 0.107 & 0.040 & 0.045 & 0.103 & 0.444 & 0.086 & 0.419 \\ \hline
PatchCore* & 0.087 & 0.043 & 0.042 & 0.407 & 0.003 & 0.443 & 0.352 & 0.189 & 0.058 & 0.017 & 0.124 & 0.028 & 0.053 & 0.270 & \underline{0.604}& 0.181 & 0.343 \\
UniAD* & 0.734 & 0.232 & \underline{0.313}& 0.517 & \underline{0.204}& 0.378 & 0.360 & 0.587 & 0.346 & 0.035 & 0.428 & \underline{0.398}& \textbf{0.542}& 0.378 & 0.443 & 0.393 & 0.086 \\ 
UCAD & \underline{0.752}& \underline{0.290}& \textbf{0.349} & \underline{0.622}& 0.187 & 0.506 & 0.333 & \underline{0.775}& \underline{0.634}& \textbf{0.214}& \underline0.549& 0.298 & 0.398 & \underline{0.535}& 0.398 & \underline{0.456}& \textbf{0.013} \\ \hline
Ours & \textbf{0.857}& \textbf{0.623}& 0.307& \textbf{0.825}& \textbf{0.259}& \textbf{0.714}& \textbf{0.547}& \textbf{0.803}& \textbf{0.690}& \underline{0.160}& \textbf{0.888}& \textbf{0.659}& \underline{0.479}& \textbf{0.686}& 0.566& \textbf{0.604}& \underline{0.040}\\ \hline
\end{tabular}
}
\end{table*}

In the text branch, the last feature $F_{T}$ from the text encoder, rich in semantic information, and the visual feature $F_{V}$ are extracted under the guidance of the task-associated text prompt. Cross-modal cosine similarity is then calculated to obtain the text-guided anomaly score $S_{T}$. To enhance the discriminative robustness of anomaly scores and better adapt to cross-task distribution differences, we designed an adaptive normalization Mechanism (ANM). This mechanism dynamically normalizes the anomaly score by adjusting the steepness and center position of the Sigmoid function through hyperparameters $k$ and $b$, optimizing the anomaly score. Given the significant computational burden of simultaneously optimizing $k$ and $b$ using a greedy search strategy, $k$ is set as a fixed value of 1.5 in this study. Experimental results demonstrate that this configuration already achieves significant improvements in the segmentation task. Further ablation studies on the choice of $k$ values will be conducted in Section \ref{sec: Ablation Study}. For hyperparameter $b$, our greedy search strategy can be described as follows: define the step set $\Delta = \{0, \pm 0.1, \pm 0.5, \pm 1, \pm 3\}$. During each training iteration in the task adaptation phase, we generate the candidate value set $ b_s =\{b_{old} + \delta | \delta \in \Delta \}$. For each candidate value $b$, the corresponding anomaly score is calculated, and the optimal candidate $b_{new}$ is selected by evaluating the detection performance on the validation set. The update rule is as follows:
\begin{equation}
    b_{\text{new}} = \arg \max_{b \in \{b_{\text{old}} + \delta\}} P(b), 
\end{equation}
where P (b) represents the performance indicators of the verification set corresponding to the current candidate b. Further, the improved normalized function is defined as:
\begin{equation}
    \sigma(x) \, = \, \frac{1}{1 + e^{-k(x - b_{\mathrm{new}})}}.
\end{equation}
Through dynamically searching the center position of the Sigmoid function, ANM adaptively compresses the score interval of normal samples and magnifies the response difference of abnormal regions, thereby achieving more accurate anomaly location.  Finally, $S_{V}$ and $S_{T}$ are up-sampled to a 224×224 resolution via bilinear interpolation, and the anomaly fraction diagrams $M_{V}$ and $M_{T}$ are obtained.  The final anomaly score map $_{M_{final}}$ is then generated using the following dynamic fusion strategy (DFS):
\begin{equation}
    M_{\text{final}} = \alpha \cdot M_V + (1 - \alpha) \cdot M_T, 
\end{equation}
among them, the hyperparameter $\alpha$ is used to balance the contribution of the two types of abnormal graphs, and $.$ denotes matrix multiplication. See Section 4.3 for the ablation experiment of $\alpha$. DSG-AFM not only makes full use of the local discriminant nature of visual features and the semantic guidance of text prompts but also alleviates the interference caused by the differences in data distribution across tasks through dynamic normalization. Ablation experiments (Table \ref{tab:7}) validated a significant improvement in AUPR indicators.
\begin{figure*} 
\centering 
\includegraphics[width=1\linewidth]{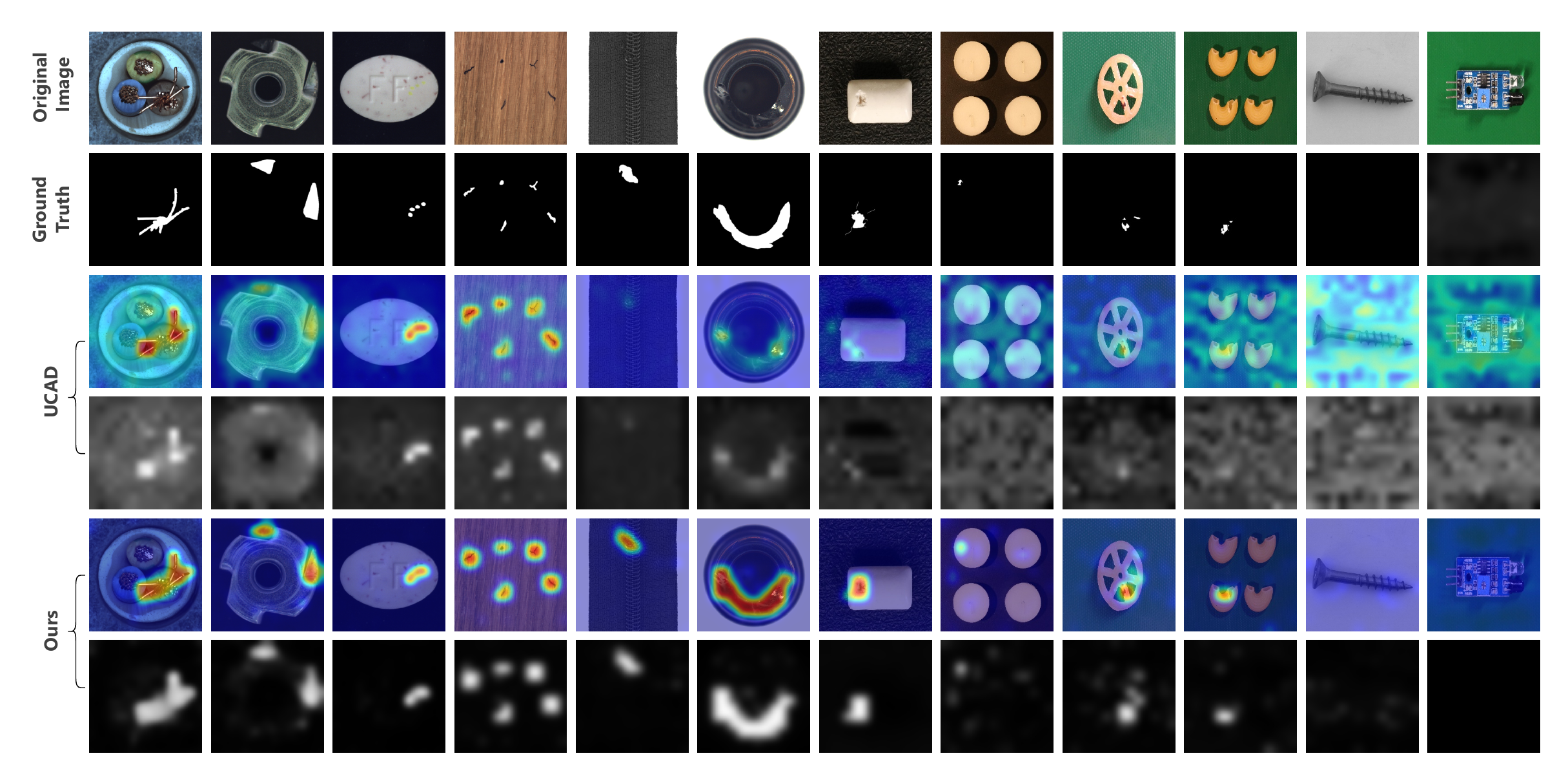} 
\Description{Visualization and analysis of the model performance}
\caption{Visualization and analysis of the model performance on the MVTec-AD and VisA Datasets.} 
\label{Fig.3} 
\end{figure*}

\begin{table*}
\centering
\caption{Image-level AUROC $\uparrow$ and corresponding FM $\downarrow$ on VisA dataset \cite{35} after training on the last subdataset.}
\label{table:3}
\resizebox{\linewidth}{!}{
\begin{tabular}{l|cccccccccccc|cc}
\hline
\textbf{Methods} & \textbf{candle} & \textbf{capsules} & \textbf{cashew} & \textbf{chewinggum} & \textbf{fryum} & \textbf{macaroni1} & \textbf{macaroni2} & \textbf{pcb1} & \textbf{pcb2} & \textbf{pcb3} & \textbf{pcb4} & \textbf{pipe\_fryum} & \textbf{average} & \textbf{avg FM} \\ \hline
CFA & 0.512 & 0.672 & 0.873 & 0.753 & 0.304 & 0.557 & 0.422 & 0.698 & 0.472 & 0.449 & 0.407 & \underline{0.998}& 0.593 & 0.327 \\ 
RD4AD & 0.380 & 0.385 & 0.737 & 0.539 & 0.533 & 0.607 & 0.487 & 0.437 & 0.672 & 0.343 & 0.187 & \textbf{0.999} & 0.525 & 0.423 \\ 
PatchCore & 0.401 & 0.605 & 0.624 & 0.907 & 0.334 & 0.538 & 0.437 & 0.527 & 0.597 & 0.507 & 0.588 & \underline{0.998}& 0.589 & 0.361 \\
SimpleNet & 0.504 & 0.474 & 0.794 & 0.721 & 0.684 & 0.567 & 0.447 & 0.598 & 0.629 & 0.538 & 0.493 & 0.945 & 0.616 & 0.283 \\ 
UniAD & 0.573 & 0.599 & 0.661 & 0.758 & 0.504 & 0.559 & 0.644 & 0.749 & 0.523 & 0.547 & 0.562 & 0.989 & 0.639 & 0.297 \\ 
DNE & 0.486 & 0.413 & 0.735 & 0.585 & 0.691 & 0.584 & 0.546 & 0.633 & 0.693 & 0.642 & 0.562 & 0.747 & 0.610 & 0.179 \\ \hline
PatchCore* & 0.647 & 0.579 & 0.669 & 0.735 & 0.431 & 0.631 & 0.624 & 0.617 & 0.534 & 0.479 & 0.645 & \textbf{0.999} & 0.633 & 0.349 \\ 
UniAD* & \underline{0.884}& 0.669 & \underline{0.938}& \textbf{0.970}& \underline{0.812}& 0.753 & 0.570 & \underline{0.872}& 0.766 & 0.708 & \textbf{0.967}& 0.990 & 0.825 & 0.125 \\ 
UCAD & 0.778 & \textbf{0.877} & \textbf{0.960} & 0.958 & \textbf{0.945}& \underline{0.823}& \underline{0.667}& \textbf{0.905}& \textbf{0.871}& \underline{0.813}& 0.901& 0.988 & \underline{0.874}& \textbf{0.039}\\ \hline
Ours & \textbf{0.944}& \underline{0.822}& 0.909& \underline{0.965}& \textbf{0.945}& \textbf{0.910} & \textbf{0.830}& 0.847& \underline{0.864}& \textbf{0.874}& \underline{0.905}& 0.997& \textbf{0.901}& \underline{0.042}\\ \hline
\end{tabular}
}
\end{table*}
\begin{table*}
\centering
\caption{Pixel-level AUPR $\uparrow$ and corresponding FM $\downarrow$ on VisA dataset \cite{35} after training on the last subdataset.}
\label{table:4}
\resizebox{\linewidth}{!}{
\begin{tabular}{l|cccccccccccc|cc}
\hline
\textbf{Methods} & \textbf{candle} & \textbf{capsules} & \textbf{cashew} & \textbf{chewinggum} & \textbf{fryum} & \textbf{macaroni1} & \textbf{macaroni2} & \textbf{pcb1} & \textbf{pcb2} & \textbf{pcb3} & \textbf{pcb4} & \textbf{pipe\_fryum} & \textbf{average} & \textbf{avg FM} \\ \hline
CFA & 0.017 & 0.005 & 0.059 & 0.243 & 0.085 & 0.001 & 0.001 & 0.013 & 0.006 & 0.008 & 0.015 & \underline{0.592}& 0.087 & 0.184 \\ 
RD4AD & 0.002 & 0.005 & 0.061 & 0.045 & 0.098 & 0.001 & 0.001 & 0.013 & 0.008 & 0.008 & 0.013 & 0.576 & 0.069 & 0.201 \\ 
PatchCore & 0.012 & 0.007 & 0.055 & 0.315 & 0.082 & 0.000 & 0.000 & 0.008 & 0.004 & 0.007 & 0.010 & 0.585 & 0.090 & 0.311 \\
SimpleNet & 0.001 & 0.004 & 0.017 & 0.007 & 0.047 & 0.000 & 0.000 & 0.013 & 0.003 & 0.004 & 0.009 & 0.058 & 0.014 & \underline{0.016}\\ 
UniAD & 0.006 & 0.013 & 0.040 & 0.185 & 0.087 & 0.002 & 0.002 & 0.015 & 0.005 & 0.015 & 0.013 & 0.576& 0.080 & 0.218 \\ \hline
PatchCore* & 0.018 & 0.010 & 0.047 & 0.202 & 0.081 & 0.003 & 0.001 & 0.008 & 0.004 & 0.008 & 0.010 & 0.443 & 0.070 & 0.327 \\ 
UniAD* & \textbf{0.132} & 0.123 & 0.378 & \underline{0.574}& \textbf{0.404} & \textbf{0.041} & \textbf{0.010}& 0.612 & 0.083 & \underline{0.266}& \underline{0.232}& 0.549 & 0.283 & 0.062 \\ 
UCAD & 0.067 & \textbf{0.437} & \underline{0.580}& 0.503 & \underline{0.334}& 0.013 & 0.003 & \textbf{0.702} & \textbf{0.136} & \underline{0.266}& 0.106 & 0.457 & \underline{0.300}& \textbf{0.015}\\ \hline
Ours & \underline{0.119}& \underline{0.351}& \textbf{0.655}& \textbf{0.730}& 0.309& \underline{0.021}& \underline{0.006}& \underline{0.689}& \underline{0.100}& \textbf{0.383}& \textbf{0.260}& \textbf{0.754}& \textbf{0.365}& 0.041\\ \hline
\end{tabular}
}
\end{table*}
\section{Experiments}
\subsection{Experimental Setup}
In this study, we conducted experiments on two widely used industrial image anomaly detection datasets: MVTec AD \cite{34} and VisA \cite{35}.MVTec AD contains 15 categories of normal and anomalous images covering industrial products such as bottles, screws, and metal parts, and provides both image-level and pixel-level annotations, which are suitable for evaluating the performance of unsupervised anomaly detection methods. VisA is currently the largest real industrial anomaly detection dataset with pixel-level annotations, contains diverse industrial scenes, and is suitable for training and evaluating models that perform both image-level and pixel-level anomaly detection.

Competing Methods and Baselines. Typical approaches from a variety of different paradigms are considered in this article to compare with our approach. These methods include CFA\cite{41}, CSFlow \cite{42}, CutPaste \cite{5}, DNE \cite{32}, DRAEM \cite{45}, Fast-Flow \cite{46}, FAVAE \cite{47}, PaDiM \cite{48}, PatchCore \cite{24}, RD4AD \cite{50}, SPADE \cite{51}, STPM \cite{52}, SimpleNet \cite{4}, UniAD \cite{6} and UCAD\cite{8}.
\begin{table}
    \centering
    \caption{Image-level AUROC and Pixel-level AUPR Performance on MVTec AD and VisA dataset with CMPMB and DSG-AFM.}
    \label{tab:5}
    \resizebox{\linewidth}{!}{ 
        \begin{tabular}{cccccc}
            \hline
            \multicolumn{2}{c}{\textbf{CMPMB}} & \multicolumn{2}{c}{\textbf{DSG-AFM}} & 
             \multirow{2}{*}{\textbf{MVtec AD}} & \multirow{2}{*}{\textbf{VisA}}  \\ \cline{1-4}
            \parbox{2cm}{\centering \textbf{Text} \\ \textbf{Prompts}}   & \parbox{2cm}{\centering \textbf{Vision} \\ \textbf{Prompts}} & \textbf{DFS} & \textbf{ANM} \\ \hline
            X                & X                & X                & X     & 0.760/0.249             & 0.808/0.127            \\
            $\checkmark$            & $\checkmark$            & $\checkmark$              & X             & 0.971/0.523             & 0.899/0.317\\
            X            & $\checkmark$            &  X             & $\checkmark$   & 0.973/0.597             & 0.891/0.347 \\
            $\checkmark$  &  $\checkmark$  & $\checkmark$  &  $\checkmark$    & \textbf{0.974/0.604}             & \textbf{0.901/0.365}\\ 
            \hline
        \end{tabular}
    }
\end{table}
To comprehensively evaluate the model performance, we use three main metrics. First, the image-level anomaly classification capability is evaluated using the area under the receiver operating characteristic curve with exact recall (AUROC). Second, the pixel-level anomaly segmentation capability is evaluated using the area under the exact recall curve (AUPR). In addition, to assess the model's capability in continuous anomaly detection scenarios, we utilize the forgetting measure (FM) \cite{chaudhry2018riemannian} to assess the model's ability to prevent catastrophic forgetting. FM measures the amount of forgetting by calculating the amount of forgetting by the model for task j after completing k tasks.
\begin{equation}
    avg \, FM = \frac{1}{k-1} \sum_{j=1}^{k-1} \max_{l \in \{1, \ldots , k-1\}} (\mathbf{T}_{l,j} - \mathbf{T}_{k,j}).
\end{equation}

In this study, we used the pre-trained multimodal model CLIP \cite{37} as the backbone network. In the prompt(prompt) training phase, we used a batch size of 8 and employed the Adam optimizer \cite{40} with a learning rate of 0.00005 and a momentum of 0.9. The training process lasted for 50 epochs. Our persistent multimodal prompting memory $M = (K, P^{T}, P^{V}, F)$ consists of an array of floating-point arrays with sizes (15, 196, 1024) floating-point array as task identity ($K$), a floating-point array of size (15, 5, 512) as learnable textual prompts ($P^{T}$), a floating-point array of size (15, 5, 768) as fine-grained visual prompts ($P^{V}$), and a floating-point array of size (15, 196, 1024) as a normal feature bank ($F$).
\subsection{Comparison With Other Methods}
We conducted comprehensive evaluations of the aforementioned 14 methods on the MVTec AD and VisA datasets. Among them, UCAD [8], as the benchmark for unsupervised continuous anomaly detection (AD), is currently the state-of-the-art (SOTA) method. PatchCore* and UniAD* are enhanced versions based on memory bank's PatchCore \cite{24} and unified paradigm's UniAD \cite{6}, respectively, designed to simulate replay-based continuous anomaly detection.

\textbf{Quantitative Analysis:} Tables \ref{table:1}-\ref{table:4} indicate that most AD methods exhibit significant performance degradation under continuous learning settings. In contrast, our approach achieves substantial advantages without employing a replay mechanism: Compared to UniAD*, our method demonstrates improvements of +7\% in image AUROC and +21.1\% in pixel AUPR on MVTec AD, and +7.6\% in image AUROC and +8.2\% in pixel AUPR on VisA. Additionally, compared to UCAD, which relies solely on visual information, our method significantly enhances discrimination ability, achieving performance improvements of +4.4\% and +14.8\% in image AUROC and pixel AUPR, respectively, on the MVTec AD dataset. On the more complex VisA dataset, we also achieved improvements of +2.7\% and +6.5\%, respectively. Notably, while the introduction of text information leads to a slight decrease in the model's anti-forgetting rate (FM), the model generally still achieves sub-optimal or even optimal results. Given the significant performance enhancements in anomaly detection and segmentation from incorporating text information, this minor decline is deemed acceptable.

\textbf{Qualitative analysis:} Fig.\ref{Fig.3} further underscores the advantages of our method in anomaly detection and segmentation. Compared with the previous SOTA method, UCAD, our approach exhibits superior detection performance. These findings highlight the framework's excellent balance between precision and robustness.
\begin{table}
\centering
\caption{Pixel-level AUPR Performance on MVTec AD and VisA dataset with different configurations of \textbf{$\alpha$}.}
\label{tab:7}
\begin{tabular}{clclc}
\hline
 \textbf{$\alpha$} && \textbf{MVTec AD}  && \textbf{VisA} \\ \hline
 0.1 && 0.592&& 0.364\\
 0.3 && 0.601&& 0.361\\
 0.5 &&  0.603&&0.364\\
 0.7 &&  0.603&&0.364\\
 0.9 && \textbf{0.604}&& \textbf{0.365}\\
\hline
\end{tabular}
\end{table}
\begin{table}
\centering
\caption{Image-level AUROC and Pixel-level AUPR Performance on MVTec AD and VisA dataset with different configurations of ViT encoder layer. }
\label{tab:8}
\begin{tabular}{clclc}
\hline
 \textbf{Layer} && \textbf{MVTec AD}  && \textbf{VisA} \\ \hline
 1 && 0.884/0.498&& 0.815/0.230\\
 3 && 0.956/0.603&& 0.881/0.361\\
 5 &&  \textbf{0.974/0.604}&&\textbf{0.901}/0.365\\
 7 &&  0.969/0.578&&\textbf{0.901/0.379}\\
 9 && 0.946/0.481&& 0.886/0.204\\
\hline
\end{tabular}
\end{table}

\subsection{Ablation Study}
\begin{figure} 
\centering 
\includegraphics[width=0.45\textwidth]{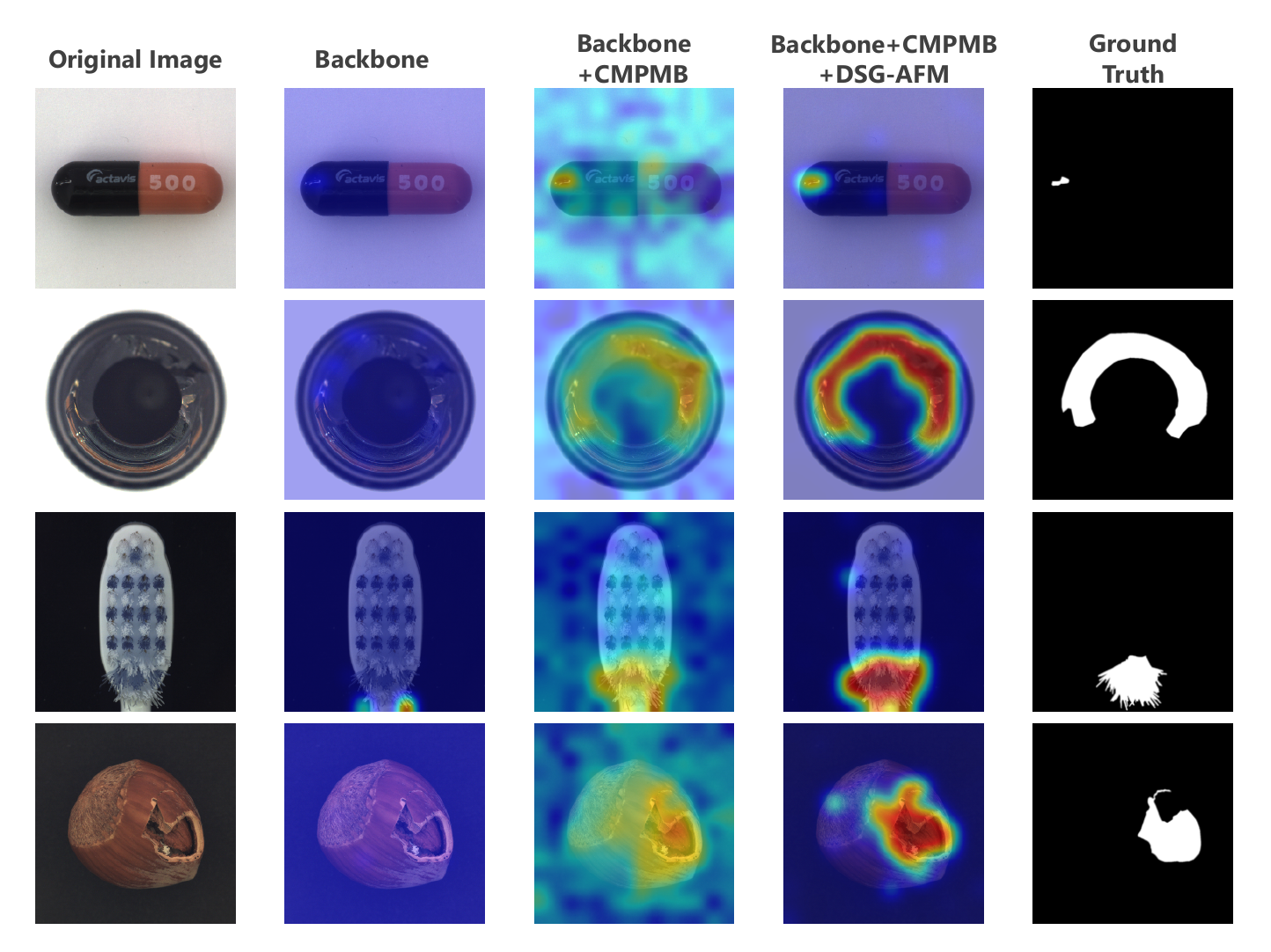} 
\Description{Visualization of anomaly segmentation under different modules on the MVTec AD dataset}
\caption{Visualization of anomaly segmentation under different modules on the MVTec AD dataset.} 
\label{Fig.4} 
\end{figure}

We conducted a series of ablation studies to further investigate the influence of each module on performance.  Initially, we performed structural ablation experiments, with results presented in Table \ref{tab:5}. In the first experiment, we removed the CMPMB and DSG-AFM modules, reverting to the most basic model configuration for anomaly detection, where the normal feature database was reset after each task. Subsequently, we introduced multimodal prompts to guide unsupervised anomaly detection in continuous scenarios, which led to significant performance improvements.  To evaluate the effectiveness of textual information, we conducted experiments using a model that relied solely on visual prompts.       The results demonstrated that removing text prompts decreased segmentation accuracy, thereby validating the importance of text prompts in continuous unsupervised anomaly segmentation tasks.   Additionally, we examined the impact of ANM on model performance and found that its removal significantly reduced abnormal segmentation performance, underscoring the critical role of ANM in segmentation tasks. The visualized results of the structural ablation experiment shown in Fig.\ref{Fig.4} further illustrate the effectiveness of CMPMB and DSG-AFM.
\label{sec: Ablation Study}
In addition, we conducted comprehensive ablation studies to investigate the impacts of critical hyperparameters on model performance.  First, we systematically evaluated the influence of the balancing coefficient $\alpha$ in our  DFS module. Table \ref{tab:7} clearly illustrates a relationship between $\alpha$ values and model performance, with optimal results achieved at $\alpha$=0.9.
Furthermore, we investigated the impact of feature extraction layers from our pre-trained visual backbone, as summarized in Table \ref{tab:8}.  Notably, the characteristics of different layers significantly impacted model performance. The experiments reveal that middle-layer features (specifically Layer 5) achieve superior performance by optimally combining detailed spatial information with high-level semantic content and lower and higher layers, leading to substantial performance declines. 
\section{Conclusion}
In this paper, we propose an unsupervised continuous anomaly detection method based on multi-modal prompts. By introducing a Continuous Multi-Modal Prompt Memory (CMPMB) and a Defect Semantics-Guided Adaptive Fusion Mechanism (DSG-AFM), this method successfully leverages multi-modal information for unsupervised persistent anomaly detection and segmentation, thereby enhancing its performance. Experimental results on the MVTec AD and VisA datasets demonstrate that our approach significantly outperforms existing unsupervised persistent anomaly detection methods, achieving state-of-the-art (SOTA) performance.    Future research could further investigate the integration of additional modal information, such as sensor data and sound, to improve the performance and generalization of the model.

\section{Acknowledgments}
\begin{acks}
This work is supported by Key R\&D Program of Shandong Province, China(2023CXGC010112),
National Natural Science Foundation of China(62401305),
the Qilu Youth Innovation Team(2024KJH028) and
Young Talent of Lifting engineering for Science and Technology in Shandong, China(SDAST2025QTA004).
\end{acks}

\bibliographystyle{ACM-Reference-Format}
\bibliography{sample-base}
\end{document}